\documentclass[11pt,a4paper]{article}
\usepackage[hyperref]{acl2021}
\usepackage{times}
\usepackage{latexsym}
\usepackage{graphicx}
\usepackage{url}
\usepackage{booktabs}
\usepackage{stfloats}
\usepackage{multirow}

\usepackage{microtype}
\aclfinalcopy

\title{fastHan: A BERT-based Multi-Task Toolkit for Chinese NLP}

\author{
Zhichao Geng, Hang Yan, Xipeng Qiu\thanks{\ \ Corresponding author} , Xuanjing Huang\\

School of Computer Science, Fudan University\\
Key Laboratory of Intelligent Information Processing, Fudan University \\
\texttt{\{zcgeng20,hyan19,xpqiu,xjhuang\}@fudan.edu.cn}

}

\date{}

\begin{document}
\maketitle
\begin{abstract}
We present fastHan, an open-source toolkit for four basic tasks in Chinese natural language processing: Chinese word segmentation (CWS), Part-of-Speech (POS) tagging, named entity recognition (NER), and dependency parsing. The backbone of fastHan is a multi-task model based on a pruned BERT, which uses the first 8 layers in BERT. We also provide a 4-layer base model compressed from the 8-layer model. The joint-model is trained and evaluated on 13 corpora of four tasks, yielding near state-of-the-art (SOTA) performance in dependency parsing and NER, achieving SOTA performance in CWS and POS. Besides, fastHan's transferability is also strong, performing much better than popular segmentation tools on a non-training corpus. To better meet the need of practical application, we allow users to use their own labeled data to further fine-tune fastHan. In addition to its small size and excellent performance, fastHan is user-friendly. Implemented as a python package, fastHan isolates users from the internal technical details and is convenient to use. The project is released on Github\footnote{\url{https://github.com/fastnlp/fastHan}}.
\end{abstract}

\section{Introduction}

Recently, the need for Chinese natural language processing (NLP) has a dramatic increase for many  downstream applications.
There are four basic tasks for Chinese NLP: Chinese word segmentation (CWS), Part-of-Speech (POS) tagging, named entity recognition (NER), and dependency parsing.
CWS is a character-level task while others are word-level tasks. These basic tasks are usually the cornerstones or provide useful features for other downstream tasks.

However, the Chinese NLP community lacks an effective toolkit utilizing the correlation between the tasks. Tools developed for a single task cannot achieve the highest accuracy, and loading tools for each task will take up more memory.
In practical, there is a strong correlation between these four basic Chinese NLP tasks. For example, the model will perform better in the other three word-level tasks if its word segmentation ability is stronger. Recently, \citet{chen2017feature} adopt cross-label to label the POS so that POS tagging and CWS can be trained jointly.
\citet{yan2020graph} propose a graph-based model for joint CWS and dependency parsing, in which a special ''APP'' dependency arc is used to indicate the word segmentation information. Thus, they can jointly train the word-level dependency parsing task and character-level CWS task with the biaffine parser~\cite{dozat2016deep}. \citet{chen2017adversarial} explore adversarial multi-criteria learning for CWS, proving more knowledge can be mined through training model on more corpora. As a result, there are many pieces of research on how to perform multi-corpus training on these tasks and how to conduct multi-task joint training.
\citet{zhang2020pos} show the joint training of POS tagging and dependency parsing can improve each other's performance and so on.  Results of the CWS task are contained in the output of the POS tagging task.

Therefore, we developed fastHan, an efficient toolkit with the help of multi-task learning and pre-trained models (PTMs)~\cite{qiu2020ptms}. FastHan adopts a BERT-based \citep{devlin2018bert} joint-model on 13 corpora to address the above four tasks. Through multi-task learning, fastHan shares knowledge among the different corpora. This shared information can improve fastHan's performance on these tasks. Besides, training on more corpora can obtain a larger vocabulary, which can reduce the number of times the model encounters characters outs of vocabulary. What's more, the joint-model can greatly reduce the occupied memory space. Compared with training a model for each task, the joint-model can reduce the occupied memory space by \textit{four} times.

FastHan has two versions of the backbone model, base and large. The large model uses the first eight layers of BERT, and the base model uses the Theseus strategy \citep{xu2020bert} to compress the large model to four layers. To improve the performance of the model, fastHan has done much optimization. For example, using the output of POS tagging to improve the performance of the dependency parsing task, using Theseus strategy  to improve the performance of the base version model, and so on.

Overall, fastHan has the following advantages:

\begin{description}
  \item[Small size:] The total parameter of the base model is 151MB, and for the large model the number is 262MB.
  \item[High accuracy:] The base version of the model achieved good results in all tasks, while the large version of the model approached SOTA in dependency parsing and NER, and achieved SOTA performance in CWS and POS.
  \item[Strong transferability:] Multi-task learning allows fastHan to adapt to multiple criteria, and a large number of corpus allows fastHan to mine knowledge from rare samples. As a result, fastHan is robust to new samples. Our experiments in section~\ref{sec:general} show fastHan outperforms popular segmentation tools on non-training dataset.

  \item[Easy to use:] FastHan is implemented as a python package, and users can get started with its basic functions in one minute. Besides, all advanced features, such as user lexicon and fine-tuning, only need one line of code to use.

\end{description}

For developers of downstream applications, they do not need to do repetitive work for basic tasks and do not need to understand complex codes like BERT. Even if users have little knowledge of deep learning, by using fastHan they can get the results of SOTA performance conveniently. Also, the smaller size can reduce the need for hardware, so that fastHan can be deployed on more platforms.

For the Chinese NLP research community, the results of fastHan can be used as a unified preprocessing standard with high quality.

Besides, the idea of fastHan is not restricted to Chinese. Applying multi-task learning to enhance NLP toolkits also has practical value in other languages.

\section{Backbone Model}

\begin{figure}
\centering
\includegraphics[width=3in]{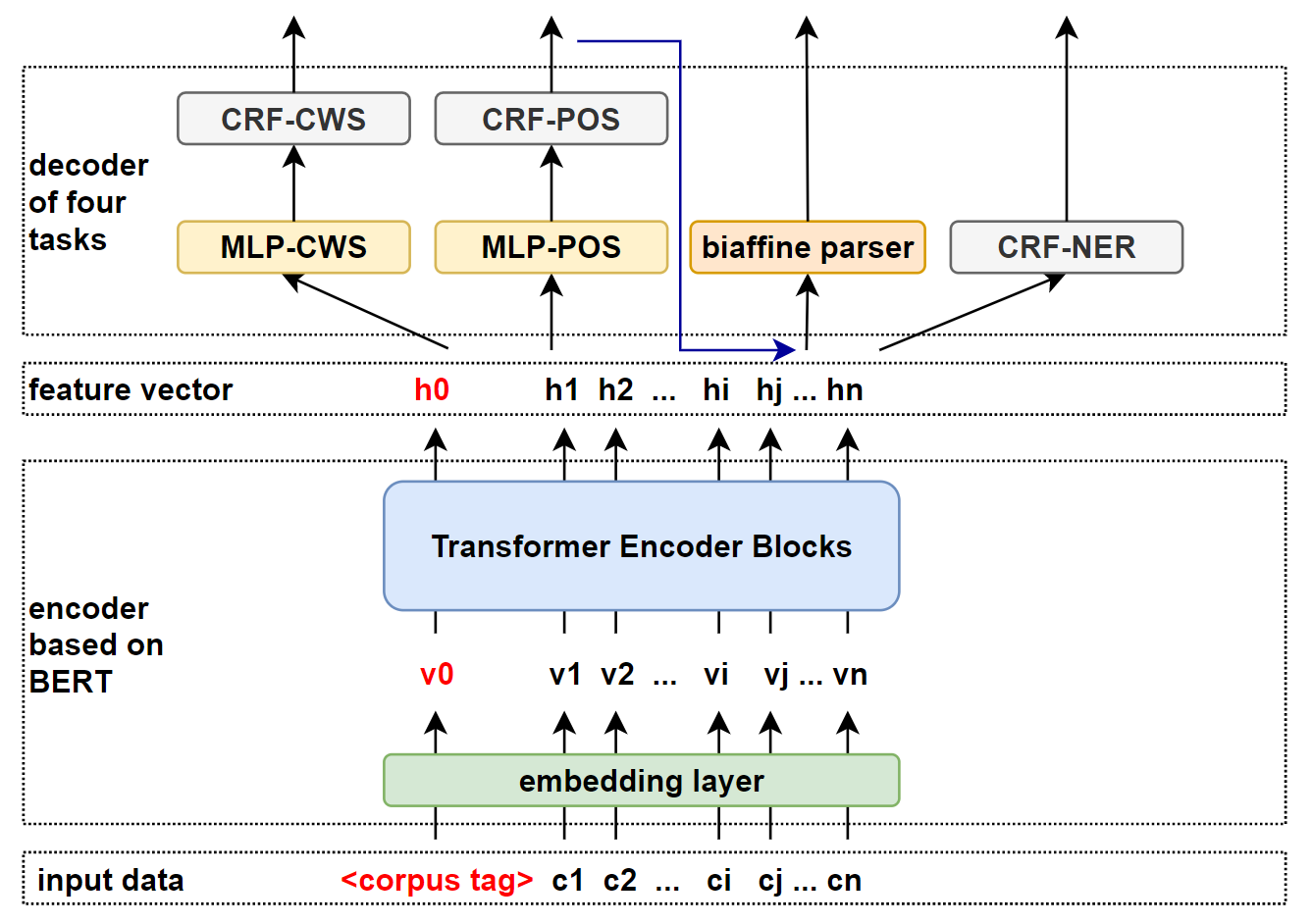}
\caption{Architecture of the proposed model. The inputs are characters embeddings.}
\label{model}
\end{figure}

The backbone of fastHan is a joint-model based on BERT, which performs multi-task learning on 13 corpora of the four tasks. The architecture of the model is shown in Figure \ref{model}. For this model, sentences of different tasks are first added with corpus tags at the beginning of the sentence. And then the sentences are input into the BERT-based encoder and the decoding layer. The decoding layer will use different decoders according to the current task: use conditional random field (CRF) to decode in the NER task; use MLP and CRF to decode in POS tagging and CWS task; use the output of POS tagging task combined with biaffine parser to decode in dependency parsing task.

Each task uses independent label sets here, CWS uses label set $Y=\{B,M,E,S\}$; POS tagging uses cross-labels set based on $\{B,M,E,S\}$; NER uses cross-labels set based on $\{B,M,E,S,O\}$; dependency parsing uses arc heads and arc labels to represent dependency grammar tree.

\subsection{BERT-based feature extraction layer}
BERT \citep{devlin2018bert} is a language model trained in large-scale corpus. The pre-trained BERT can be used to encode the input sequence. We take the output of the last layer of transformer blocks as the feature vector of the sequence. The attention \citep{vaswani2017attention} mechanism of BERT can extract rich and semantic information related to the context. In addition, the calculation of attention is parallel in the entire sequence, which is faster than the feature extraction layer based on LSTM. Different from vanilla BERT, we prune its layers and add corpus tags to input sequences.

\paragraph{Layer Pruning:}The original BERT has 12 layers of transformer blocks, which will occupy a lot of memory space. The time cost of calculating for 12 layers is too much for these basic tasks even if data flows in parallel. Inspired by \citet{huang2019toward}, we only use 4 or 8 layers. Our experiment found that using the first eight layers performs well on all tasks, and after compressing, four layers are enough for CWS, POS tagging, and NER.

\paragraph{Corpus Tags:}Instead of a linear projection layer, we use corpus tags to distinguish various tasks and corpora. Each corpus of each task corresponds to a specific corpus tag, and the embedding of these tags needs to be initialized and optimized during training. As shown in Figure \ref{model}, before inputting the sequence into BERT, we add the corpus tag to the head of the sequence. The attention mechanism will ensure that the vector of the corpus tag and the vector of each other position generate sufficiently complex calculations to bring the corpus and task information to each character.

\subsection{CRF Decoder}
\label{sec:crf}
We use the conditional random field (CRF)  \citep{lafferty2001conditional} to do the final decoding work in POS tagging, CWS, and NER tasks. In CRF, the conditional probability of a label sequence can be formalized as:
$$
P(Y|X)=\frac{1}{Z(x;\theta)}exp(\sum_{t=1}^{T}\theta_1^\top f_{1}(X,y_{t})+
$$
$$
\sum_{t=1}^{T-1}\theta_{2}^\top f_{2}(X,y_{t},y_{t+1}))~~~~~~(1)
$$
where $\theta$ are model parameters, $f_{1}(X,y_{t})$ is the score for label $y_{t}$ at position $t$, $f_{2}(X,y_{t},y_{t+1})$ is the transition score from $y_{t}$ to $y_{t+1}$, and $Z(x;\theta)$ is the normalization factor.

Compared with decoding using MLP only, CRF utilizes the neighbor information. When decoding using the Viterbi algorithm, CRF can get the global optimal solution instead of the label with the highest score for each position.

\subsection{Biaffine Parser with Output of POS tagging}
This task refers to the work of \citet{yan2020graph}. Yan's work uses the biaffine parser to address both CWS and dependency parsing tasks. Compared with the work of \citet{yan2020graph}, our model will use the output of POS tagging for two reasons. First, dependency parsing has a large semantic and formal gap with other tasks. As a result, sharing the parameter space with other tasks will reduce its performance. Our experimental results show that when the prediction of dependency parsing is independent of other tasks, the performance is worse than that of training dependency parsing only. And using the output of POS, dependency parsing can get more useful information, such as word segmentation and POS tagging labels. More importantly, users have the need to obtain all information in one sentence. If running POS tagging and dependency parsing separately, the word segmentation results of the two tasks may conflict, and this contradiction cannot be resolved by engineering methods. Even if there is error propagation in this way, our experiment shows the negative impact is acceptable with high POS tagging accuracy.

When predicting for dependency parsing, we first add the POS tagging corpus tag at the head of the original sentence to get the POS tagging output. Then we add the corpus tag of dependency parsing at the head of the original sentence to get the feature vector. Then, using the word segmentation results from POS tagging to split the feature vector of dependency parsing by token. The feature vectors of characters in a token are averaged to represent the token. In addition, embedding is established for POS tagging labels, with the same dimension as the feature vector. The feature vector of each token is added to the embedding vector by position, and the result is input into the biaffine parser. During the training phase, the model uses golden POS tagging labels. The premise of using POS tagging output is that the corpus contains both dependency parsing and POS tagging information.

\subsection{Theseus Strategy}

\begin{figure}

\centering
\includegraphics[width=3in]{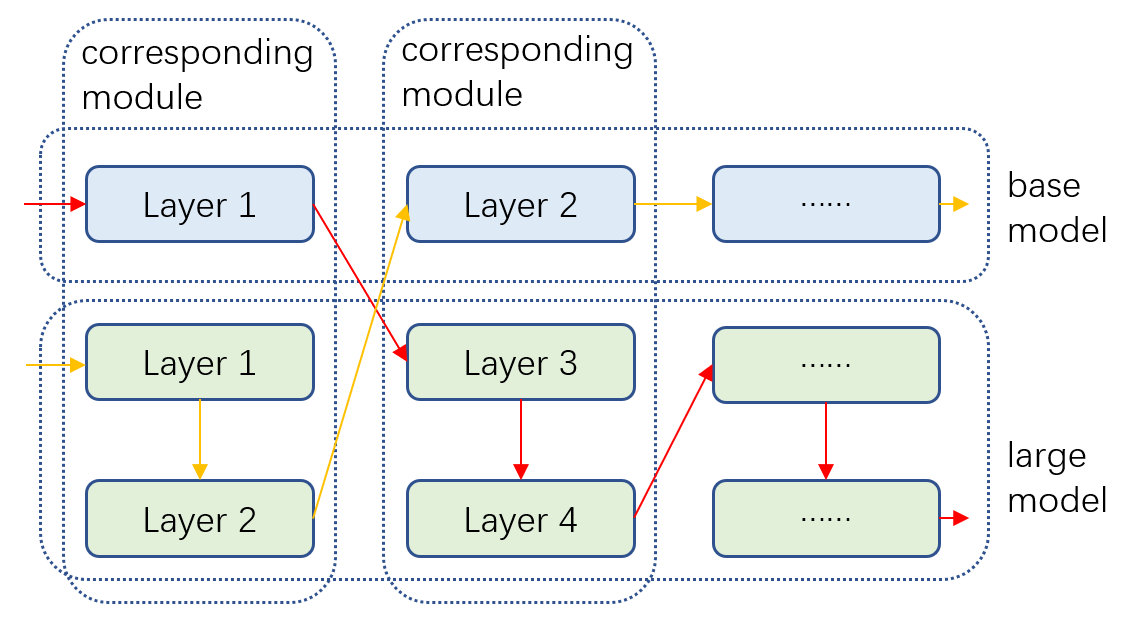}
\caption{This diagram explains the replacement strategy when using Theseus method. When training the base model, we randomly replace the layer of base model with corresponding layers of large model. The red arrows and yellow arrows represent two possible data paths during training.}
\label{theseus}
\end{figure}

Theseus strategy \citep{xu2020bert} is a method to compress BERT, and we use it to train the base version of the model. As shown in Figure \ref{theseus}, after getting the large version of the model we use the module replacement strategy to train the four-layer base model. The base model is initialized with the first four layers of the large model, and its layer $i$ is bound to the layer $2i-1$ and $2i$ of the large model. They are the corresponding modules. The training phase is divided into two parts. In the first part, we randomly choose whether to replace the module in the base model with its corresponding module in the large model. And we make the choice for each module. We freeze the parameters of the large model when using gradients to update parameters. The replacement probability $p$ is initialized to 0.5 and decreases linearly to 0. In the second part, We only fine-tune the base model and don't replace the modules anymore.

\begin{figure}

\centering
\includegraphics[width=3in]{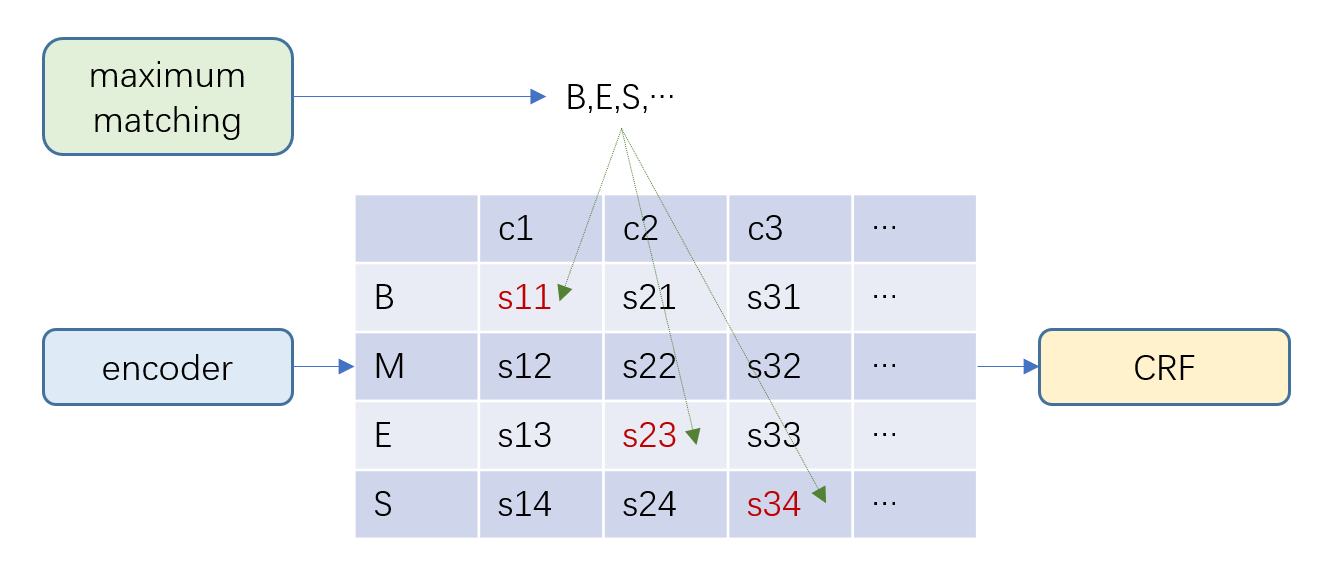}
\caption{An example of segmentation of sequence $(c1,c2,c3,...)$ combined with a user lexicon. According to the segmentation result of the maximum matching algorithm, a bias will be added to scores marked in red. }
\label{lexicon}
\end{figure}

\subsection{User Lexicon}

In actual applications, users may process text of specific domains, such as technology, medical. There are proprietary vocabularies with high recall rates in such domains, and they rarely appear in ordinary corpus. It is intuitive to use a user lexicon to address this problem. Users can choose whether to add or use their lexicon. An example of combining a user lexicon is shown in Figure \ref{lexicon}. When combined with a user lexicon, the maximum matching algorithm \citep{10.3115/992628.992665} is first performed to obtain a label sequence. After that, a bias will be added to the corresponding scores output by the encoder. And the result will be viewed as $f_{1}(X,y_{t})$ in CRF in section~\ref{sec:crf}. The bias is calculated by the following equation:
$$
b_{t}=(max(y_{1:n})-average(y_{1:n}))*w~~~~~~(2)
$$
where $b_{t}$ is the bias on position t, $y_{1:n}$ is the scores of each labels on position t output by the encoder, and $w$ is the coefficient whose default value is 0.05. CRF decoder will generate the global optimal solution considering the bias. Users can set the coefficient value according to the recall rate of their lexicon. A development set can also be applied to get the optimal coefficient.

\section{fastHan}

FastHan is a Chinese NLP toolkit based on the above model, developed based on fastNLP\footnote{\url{https://github.com/fastnlp/fastnlp}} and PyTorch. We made a short video demonstrating fastHan and uploaded it to YouTube\footnote{\url{https://youtu.be/apM78cG06jY}} and bilibili\footnote{\url{https://www.bilibili.com/video/BV1ho4y117H3}}.

FastHan has been released on PYPI and users can install it by pip:
\begin{quote}
\small
\begin{verbatim}
pip install fastHan
\end{verbatim}
\end{quote}

\begin{figure}

\centering
\includegraphics[width=3in]{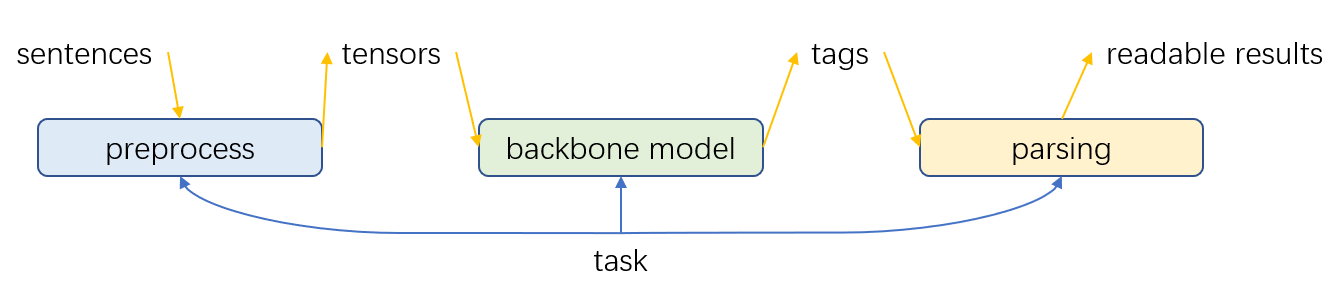}
\caption{The workflow of fastHan. As indicated by the yellow arrows, data is converted between various formats in each stage. The blue arrows reveal that fastHan needs to act according to the task being performed currently. }
\label{workflow}
\end{figure}

\begin{figure*}[t]

\centering
\includegraphics[width=6in]{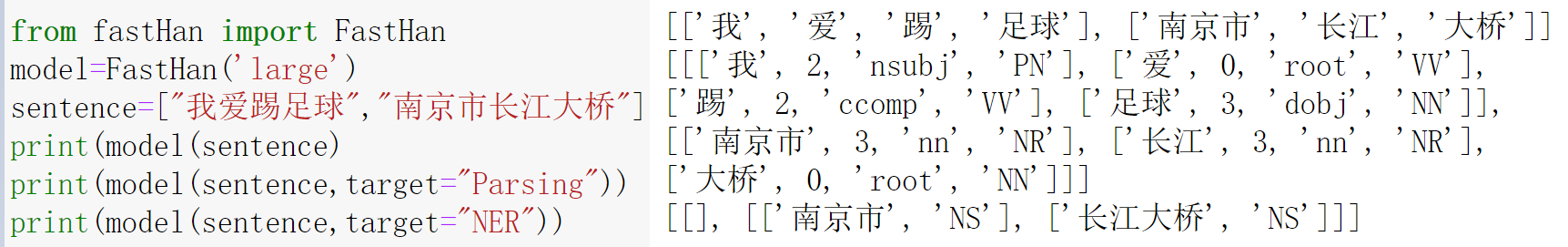}
\caption{An example of using fastHan. On the left is the code entered by the user, and on the right is the corresponding output. The two sentences in the figure mean "I like playing football" and "Nanjing Yangtze River Bridge". The second sentence can be explained in a second way as "Daqiao Jiang, mayor of the Nanjing city", and it is quite easy to include a user lexicon to customize the output of the second sentence.}
\label{usage}
\end{figure*}

\subsection{Workflow}
When FastHan initializes, it first loads the pre-trained model parameters from the file system. Then, fastHan uses the pre-trained parameters to initialize the backbone model. FastHan will download parameters from our server automatically if it has not been initialized in the current environment before. After initialization, FastHan's workflow is shown in Figure \ref{workflow}.

In the preprocessing stage, fastHan first adds a corpus tag to the head of each sentence according to the current task and then uses the vocabulary to convert the sentence into a batch of vectors as well as padding. FastHan is robust and does not preprocess the original sentence redundantly, such as removing stop words, processing numbers and English characters.

 In the parsing phase, fastHan first converts the label sequence into character form and then parses it. FastHan will return the result in a form which is readable for users.

\subsection{Usage}
As shown in Figure \ref{usage}, fastHan is easy to use. It only needs one line of code to initialize, where users can choose to use the base or large version of the model.

When calling fastHan, users need to select the task to be performed. The information of the three tasks of CWS, POS, and dependency parsing is in an inclusive relationship. And the information of the NER task is independent of other tasks. The input of FastHan can be a string or a list of strings. In the output of fastHan, words and their attributes are organized in the form of a list, which is convenient for subsequent processing. By setting parameters, users can also put their user lexicon into use. FastHan uses CTB label sets for POS tagging and dependency parsing tasks, and uses MSRA label set for NER.

Besides, users can call the $set\_device$ function to change the device utilized by the backbone model. Using GPU can greatly accelerate the prediction and fine-tuning of fastHan.

\subsection{Advanced Features}

In addition to using fastHan as a off the shelf model, users can utilize user lexicon and fine-tuning to enhance the performance of fastHan. As for user lexicon, users can call the $add\_user\_dict$ function to add their lexicon, and call the $set\_user\_dict\_weight$ function to change the weight coefficient. As for fine-tuning, users can call the $finetune$ function to load the formatted data, make fine-tuning, and save the model parameters.

Users can change the segmentation style by calling the $set\_cws\_style$ function. Each CWS corpus has different granularity and coverage. By changing the corpus tag, fastHan will segment words in the style of the corresponding corpus.

\section{Evaluation}

We evaluate fastHan in terms of accuracy, transferability, and execution speed.

\subsection{Accuracy Test}
The accuracy test is performed on the test set of training data. We refer to the CWS corpora used by \cite{chen2015long,huang2019toward}, including PKU, MSR, AS, CITYU \citep{emerson2005second}, CTB-6 \citep{xue2005penn}, SXU \citep{jin2008fourth}, UD, CNC, WTB \citep{wang2014dependency} and ZX \citep{zhang2014type}. More details can be found in \citep{huang2019toward}. For POS tagging and dependency parsing, we use the Penn Chinese Treebank 9.0 (CTB-9) \citep{xue2005penn}. For NER, we use MSRA's NER dataset and OntoNotes.

We conduct an additional set of experiments to make the base version of fastHan trained on each task separately. The final results are shown in Table \ref{tab:res}. Both base and large models perform satisfactorily. The result shows that multi-task learning greatly improves fastHan's performance on all tasks. The large version of fastHan outperforms the current best model in CWS and POS.  Although fastHan's score on NER and dependency parsing is not the best, the parameters used by fastHan are reduced by one-third due to layer prune. FastHan's performance on NER can also be enhanced by a user lexicon with a high recall rate.

We also conduct an experiment about user lexicon on 10 CWS corpus respectively. With each corpus, a word is added to the lexicon once it has appeared in the training set. With such a low-quality lexicon, fastHan's score increases by an average of 0.127 percentage points. It is feasible to use user lexicon to enhance fastHan's performance in specific domains.

 \begin{table*}[htb]
\small
  \centering
    \begin{tabular}{cccccc}
      \toprule
      \textbf{Model} & \textbf{CWS} & \textbf{ Dependency Parsing}&\textbf{POS}& \textbf{NER MSRA}& \textbf{NER OntoNotes}\\

       & $F$ & $F_{udep}$, $F_{ldep}$& $F$& $F$& $F$ \\

      \midrule

      SOTA models & 97.1 & \textbf{85.66}, \textbf{81.71} & 93.15 & \textbf{96.09} & 81.82\\

      \midrule

      fastHan base trained separately & 97.15 & 80.2, 75.12 & 94.27 & 92.2 & 80.3 \\

      fastHan base trained jointly& 97.27 & 81.22, 76.71 & 94.88 & 94.33 & 82.86\\

      fastHan large trained jointly& \textbf{97.41} & 85.52, 81.38 & \textbf{95.66} & 95.50 & \textbf{83.82}\\

      \bottomrule

    \end{tabular}
\caption{The results of fastHan's accuracy result. The score of CWS is the average of 10 corpora. When training dependency parsing separately, the biaffine parser use the same architecture as \citet{yan2020graph}. SOTA models are best-performing work we know for each task. They came from \citet{huang2019toward}, \citet{yan2020graph}, \citet{meng2019glyce}, \citet{li-etal-2020-flat} in order. \citet{li-etal-2020-flat}
uses lexicon to enhance the model.}
\label{tab:res}
\end{table*}

\subsection{Transferability Test}
\label{sec:general}

\begin{table}[ht!]
\centering\small
\begin{tabular}{lc}
\toprule
\textbf{Segmentation Tool} & \textbf{Weibo Test Set}\\
\midrule
\verb|jieba| & {83.58} \\
\verb|SnowNLP| &  {79.65} \\
\verb|THULAC| &  {86.65} \\
\verb|LTP-4.0| &  {92.05} \\
\verb|fastHan| &  \textbf{93.38}  \\
\verb|fastHan(fine-tuned)| & \textbf{96.64} \\
\bottomrule
\end{tabular}
\caption{Transfer test for fastHan, using span F metric. We use the test set of Weibo, which has 8092 samples. For LTP-4.0, we use the base version, which has the best performance among their models.}
\label{tab:general}
\end{table}

For an NLP toolkit designed for the open domain, the ability of processing samples not in the training corpus is very important. We perform the transfer test on Weibo \citep{qiu2016overview}, which has no overlap with our training data. Samples in Weibo\footnote{\url{https://github.com/FudanNLP/NLPCC-WordSeg-Weibo}} come from the Internet, and they are complex enough to test the model's transferability. We choose to test on CWS because nearly all Chinese NLP tools have this feature. We choose popular toolkits as the contrast, including Jieba\footnote{\url{https://github.com/fxsjy/jieba}}, THULAC\footnote{\url{https://github.com/thunlp/THULAC}}, SnowNLP\footnote{\url{https://github.com/isnowfy/snownlp}} and LTP-4.0\footnote{\url{https://github.com/HIT-SCIR/ltp}}. We also perform a test of fine-tuning using the training set of Weibo.

The results are shown in Table \ref{tab:general}. As a off the shelf model, FastHan outperforms  jieba, SnowNLP, and THULAC a lot. LTP-4.0 \citep{che2020n} is another technical route for multi-task Chinese NLP, which is released after the first release of fastHan. However, FastHan still outperforms LTP with a much smaller model (262MB versus 492MB). The result proves fastHan is robust to new samples, and the fine-tuning feature allows fastHan to better be adapted to new criteria.

\subsection{Speed Test}

\begin{table}[ht!]
\centering\small
\begin{tabular}{lcc}
\toprule
\multirow{2}*{\textbf{Models}} &\textbf{Dependency Parsing}&\textbf{Other Tasks}\\
& CPU, GPU & CPU, GPU\\
\midrule
\verb|fastHan base| & 25, 22 & 55, 111 \\
\verb|fastHan large| & 14, 21 & 28, 97 \\
\bottomrule
\end{tabular}
\caption{Speed test for fastHan. The numbers in the table represent the average number of sentences processed per second.}
\label{tab:speed}
\end{table}

The speed test was performed on a personal computer configured with Intel Core i5-9400f + NVIDIA GeForce GTX 1660ti. The test was conducted on the first 800 sentences of the CTB CWS corpus, with an average of 45.2 characters per sentence and a batch size of 8.

The results are shown in Table \ref{tab:speed}. Dependency parsing runs slower, and the other tasks run at about the same speed. The base model with GPU performs poorly in dependency parsing because dependency parsing requires a lot of CPU calculations, and the acceleration effect of GPU is less than the burden of information transfer.

\section{Conclusion}
In this paper, we presented fastHan, a BERT-based toolkit for CWS, NER, POS, and dependency parsing in Chinese NLP. After our optimization, fastHan has the characteristics of high accuracy, small size, strong transferability, and ease of use.

In the future, we will continue to improve the fastHan with better performance, more features and more efficient learning methods, such as meta-learning~\citep{ke-etal-2021-pre}.

\section*{Acknowledgements}
This work was supported by the National Key Research and Development Program of China (No. 2020AAA0106700), National Natural Science Foundation of China (No. 62022027) and Major Scientific Research Project of Zhejiang Lab (No. 2019KD0AD01).

\bibliographystyle{acl_natbib}
\bibliography{anthology,acl2021}

\end{document}